\documentclass[11pt,conference]{IEEEtran}
\IEEEoverridecommandlockouts
\usepackage{cite}
\usepackage{amsmath,amssymb,amsfonts}
\usepackage{algorithmic}
\usepackage{graphicx}
\usepackage{textcomp}

\usepackage{algorithm}

\usepackage{xcolor}
\def\BibTeX{{\rm B\kern-.05em{\sc i\kern-.025em b}\kern-.08em
    T\kern-.1667em\lower.7ex\hbox{E}\kern-.125emX}}

\title{Context-Aware Mobile Network Performance Prediction Using Network \& Remote Sensing Data}

\author{\IEEEauthorblockN{Ali Shibli, 
Tahar Zanouda}\
\IEEEauthorblockA{Global AI Accelerator, Ericsson, Sweden}\
\texttt{\{ali.a.shibli, tahar.zanouda\}@ericsson.com}}

\begin{document}

\maketitle

\begin{abstract}

Accurate estimation of Network Performance is crucial for several tasks in telecom networks. Telecom networks regularly serve a vast number of radio nodes. Each radio node provides services to end-users in the associated coverage areas. The task of predicting Network Performance for telecom networks necessitates considering complex spatio-temporal interactions and incorporating geospatial information where the radio nodes are deployed. Instead of relying on historical data alone, our approach augments network historical performance datasets with satellite imagery data. Our comprehensive experiments, using real-world data collected from multiple different regions of an operational network, show that the model is robust and can generalize across different scenarios. The results indicate that the model, utilizing satellite imagery, performs very well across the tested regions. Additionally, the model demonstrates a robust approach to the cold-start problem, offering a promising alternative for initial performance estimation in newly deployed sites.\end{abstract}


\section{Introduction}

Mobile operators across the globe are rolling out the fifth generation (5G) of networks to unlock a wide range of new services and meet high requirements for applications like vehicle-to-vehicle communication. Before rolling out 5G technologies, mobile operators assess the infrastructures and the urban environment to study the feasibility of deploying a new 5G site and upgrading an existing one. Therefore, assisting network planners with a method that predicts network performance while capturing the evolving urban infrastructure is important.

In 5G-and-beyond networks, the process can be even more challenging because of the use of high-frequency bands. High signal frequencies undergo significant attenuation and are distorted while interacting with various objects in the environment. Consequently, it makes it hard to characterize network performance without considering geospatial feature attributes in coverage areas.

Additionally, the number of mobile and IoT devices continues to soar, which forces network operators to work on the dual task of enhancing their existing 4G infrastructure and transitioning to the cutting-edge 5G technologies. One of the primary challenges in this evolution is network densification, characterized by a surge in new cell sites. This transition, though promising in terms of connectivity, brings with it the task of network planning. Network planners need to meticulously determine where to place base stations to optimize both coverage and costs. The dynamics of rapidly growing cities and expanding urban landscapes further complicate this task.

Historically, cellular network planning has been grounded in network data \cite{3GPP}, which often overlooks the intricate interplay of various factors influencing network performance. Recognizing this limitation, our research takes another approach; harnessing the power of satellite imagery to estimate Radio Access Network (RAN) performance. By analyzing satellite imagery data, we aim to discern the infrastructure elements and geographical attributes that can impact network performance. Augmenting this information to the historical RAN performance can provide a holistic understanding of network behavior.

It is often said: "We shape our buildings and afterward our buildings shape us." Our research is inspired by the idea that spatial structure inherits how end-users use and interact with telecom network. By leveraging satellite imagery, we present an approach to predict network performance in specific areas based on their structural attributes. This method relies on profiling and segmenting geographic regions that share similar underlying dynamics. These satellite images provide a unique perspective on the geographical and environmental factors that can affect network performance, such as topography, urban density, and natural foliage, which in turn influence signal coverage, capacity demands, and service quality. Furthermore, this method not only reduces computational overhead and streamlines processes but also paves the way for building data-driven digital twins for networks \cite{khan2022digital}. In essence, we are introducing a Remote Sensing-aided approach to predict RAN performance.

The main contributions of this work are four folds:

    \begin{enumerate}
        \item A forecasting model integrating geospatial data from satellite imagery to predict network KPIs effectively.

        \item A network profiling method clustering network nodes into groups by geospatial attributes, optimizing computational efficiency.

        \item Benchmarking three state-of-the-art computer vision models using the EuroSAT dataset \cite{helber2019eurosat}.
        \item A demonstration of the model’s effectiveness in addressing the cold-start problem, thereby enhancing performance estimations for new and planned network sites.
    \end{enumerate}

%Overall, this work contributes to a research field that leverages remote sensing data for telecom networks.

\section{Related work}
We outline several related works that either inspire the work of predicting Network KPIs or sit more broadly in the space of Remote Sensing for Telecom applications.

Diagnosing network performance is a problem that has historically been studied under various technologies using Network Key Performance Indicators \cite{mourad2020towards}. Machine Learning techniques have been employed to predict network performance. Tran et al. \cite{tran2023ml} proposed an LSTM model to estimate throughput in 5G and B5G networks. Nabi et al. \cite{nabi2023deep} explore model fusion techniques to combine different deep learning algorithms, namely Long Short-Term Memory (LSTM), Bidirectional LSTM (BiLSTM) and Gated Recurrent Unit (GRU), for better collective performance. Yaqoob et al. \cite{yaqoob2022data} developed a GNN-based model to predict network performance by leveraging the spatio\-temporal setup of telecom networks, where the network is modeled as a graph structure over which each node maintains a time series. Moreover,  KPI prediction has been used for fault detection in mobile network \cite{FaultDetectionLSTM} \cite{FaultDetectionGNN} where the predicted performance is compared with real value to detect anomalous data points. 

In recent years, there have been a number of papers exploring the interplay between remote sensing and telecom. Thrane et al. \cite{thrane2020deep} presented a model-aided deep learning approach for path loss prediction. In their work, they augmented radio data with rich and unconventional information about the site, e.g. satellite photos, to provide more accurate and flexible models. Similarly, Zhang et al. \cite{zhang2020cellular} used top-view geographical images for Radio propagation modeling.

\section{Data}
\subsection{Key Performance Indicators (KPIs) data}
\label{sec:pm-data}

Mobile operators monitor the network continuously to track the behavior of the network using \textit{Performance Management (PM) data} to gauge network performance. PM data is captured at regular intervals across Radio nodes, in different software and hardware components. 

Network performance is assessed using \textit{Key Performance Indicators (KPIs)} \cite{3GPP}. KPIs are a set of formulas to calculate Performance Indicators using PM \cite{3GPP} and CM \cite{3GPP} data. KPIs are standardized by \textit{3GPP}. There exist different categories of \textit{KPI}s, such as accessibility, mobility, integrity, utilization, and energy performance. The behavior of these \textit{KPI}s can vary depending on service area characteristics.

KPIs possess a temporal dimension, reflecting the dynamic and evolving nature of network performance over time. This temporal aspect is crucial for understanding patterns, trends, and anomalies in network behavior, as the performance indicators captured at various radio nodes are not static but fluctuate based on numerous factors such as network load, user behavior, and environmental conditions. The temporal variability makes KPIs data a perfect fit for time series analysis techniques to model and predict network performance. 
\newline
\textbf{KPIs Data Pre-Processing}: PM data was collected over 2 months for different regions of interest, for a total of over 2000 nodes from different cities and regions, using 80\% of the data for training, and 20\% of the data for testing. PM data is aggregated over 15 minutes. We normalize this data using Min-Max normalization over the 2 months. 24 hours of historical data points are used to forecast the future 8 hours, resulting in a history size of 96 data points and a horizon size of 32 data points.

\subsection{Cellular coverage areas}
\label{sec:cellular-coverage-areas}

Telecom networks consist of a set of interconnected nodes located on physical sites, each cell in a site provides coverage in a geographic area. Each cellular site or antenna is aimed at a specific geographical region. This targeted region or sector can be deduced using a combination of data points like Cell Latitude, Cell Longitude, Cell Azimuth, Antenna Tilt, and Cell Range. We approximate this sector shape as a rectangle, effectively representing each coverage area. A sector area can be estimated as described in this method \cite{coverage_area}.

\subsection{Satellite Imagery}
\label{sec:satellite-imagery}

Satellite imagery plays a pivotal role in our approach, offering unique insights into the geographical and environmental contexts that influence network performance. Satellites operated by both governmental and private sectors, address numerous needs, such as agriculture, climate monitoring, military reconnaissance, and urban landscape observation. One of the available satellites, the Sentinel satellites series \cite{sentinel2} are frequently utilized in commercial and research settings due to their easy accessibility, cost-effectiveness (being open source), and extensive documentation. The Sentinel series offers up to 10-meter spatial resolution for the RGB bands. Thus, for each region of interest, we query satellite imagery and extract the RGB bands to form our images for training the models. Finally, we use the coverage areas defined in \ref{sec:cellular-coverage-areas} to extract the satellite imagery for those coverage areas for each cell in the associated networks. An important note is that we use data from the same season over the regions of interest (Spring), since we assume that weather can affect the network performance.

% Our experiments suggest that other bands can also yield insightful results, expanding the potential for diverse applications. The precision and detail of such images can be seen in Figure X.X, depicting the Geneva region.

% [Include the figure code here]

\subsection{EuroSAT}
\label{sec:eurosat}

EuroSAT dataset \cite{helber2019eurosat} provides a diverse collection of images sourced from the Sentinel-2 satellite. With a spectrum spanning 13 bands, the dataset houses 27,000 labeled images, each structured as a 64x64 pixel grid, and belongs to one of ten different classes \textit{(Annual Crop, Forest, Herbaceous Vegetation, Highway, Industrial, Pasture, Permanent Crop, Residential, River, Sea and Lake)}. Each image in the dataset possesses a resolution that ranges between 10m, 20m, and 60m, depending on the specific band.

For our study, we use the RGB bands from the EuroSAT dataset, aligning with the spectral characteristics of the satellite images we gathered for the regions of interest. The EuroSAT images are used to fine-tune the computer vision model to be used later for representation embedding: 70\% of the images for training and 30\% for testing.

\bibliographystyle{ieeetr}
\bibliography{bibliography}

\begin{thebibliography}{10}

\bibitem{3GPP}
3GPP, ``The 3rd generation partnership project,'' tech. rep., 2023.

\bibitem{khan2022digital}
L.~U. Khan, Z.~Han, W.~Saad, E.~Hossain, M.~Guizani, and C.~S. Hong, ``Digital
  twin of wireless systems: Overview, taxonomy, challenges, and
  opportunities,'' {\em IEEE Communications Surveys \& Tutorials}, 2022.

\bibitem{helber2019eurosat}
P.~Helber, B.~Bischke, A.~Dengel, and D.~Borth, ``Eurosat: A novel dataset and
  deep learning benchmark for land use and land cover classification,'' 2019.

\bibitem{mourad2020towards}
A.~Mourad, R.~Yang, P.~H. Lehne, and A.~De~La~Oliva, ``Towards 6g: Evolution of
  key performance indicators and technology trends,'' in {\em 2020 2nd 6G
  wireless summit (6G SUMMIT)}, pp.~1--5, IEEE, 2020.

\bibitem{tran2023ml}
N.~P. Tran, O.~Delgado, B.~Jaumard, and F.~Bishay, ``Ml kpi prediction in 5g
  and b5g networks,'' in {\em 2023 Joint European Conference on Networks and
  Communications \& 6G Summit (EuCNC/6G Summit)}, pp.~502--507, IEEE, 2023.

\bibitem{nabi2023deep}
S.~T. Nabi, M.~R. Islam, M.~G.~R. Alam, M.~M. Hassan, S.~A. AlQahtani, G.~Aloi,
  and G.~Fortino, ``Deep learning based fusion model for multivariate lte
  traffic forecasting and optimized radio parameter estimation,'' {\em IEEE
  Access}, vol.~11, pp.~14533--14549, 2023.

\bibitem{yaqoob2022data}
M.~Yaqoob, R.~Trestian, and H.~X. Nguyen, ``Data-driven network performance
  prediction for b5g networks: a graph neural network approach,'' in {\em 2022
  IEEE Ninth International Conference on Communications and Electronics
  (ICCE)}, IEEE, 2022.

\bibitem{FaultDetectionLSTM}
T.~Zanouda, S.~Govindaraj, D.~Budyn, and M.~Rydar, ``Methods and apparatuses
  for detecting and localizing faults using machine learning models,'' 2022.
\newblock PCT/EP2022/071544.

\bibitem{FaultDetectionGNN}
R.~Bourgerie and T.~Zanouda, ``Fault detection in telecom networks using
  bi-level federated graph neural networks,'' in {\em 2023 IEEE International
  Conference on Data Mining Workshops (ICDMW)}, pp.~1608--1617, 2023.

\bibitem{thrane2020deep}
J.~Thrane, B.~Sliwa, C.~Wietfeld, and H.~L. Christiansen, ``Deep learning-based
  signal strength prediction using geographical images and expert knowledge,''
  in {\em GLOBECOM 2020-2020 IEEE Global Communications Conference}, pp.~1--6,
  IEEE, 2020.

\bibitem{zhang2020cellular}
X.~Zhang, X.~Shu, B.~Zhang, J.~Ren, L.~Zhou, and X.~Chen, ``Cellular network
  radio propagation modeling with deep convolutional neural networks,'' in {\em
  Proceedings of the 26th ACM SIGKDD International Conference on knowledge
  discovery \& data mining}, pp.~2378--2386, 2020.

\bibitem{coverage_area}
T.~Zanouda, ``Methods and nodes for predicting azimuth values of cells in
  communications networks,'' 2022.
\newblock PCT/EP2022/072279.

\bibitem{sentinel2}
E.~S. Agency, ``Sentinel-2,'' 2015.
\newblock Accessed on February 2, 2024.

\bibitem{he2015deep}
K.~He, X.~Zhang, S.~Ren, and J.~Sun, ``Deep residual learning for image
  recognition,'' 2015.

\bibitem{tan2020efficientnet}
M.~Tan and Q.~V. Le, ``Efficientnet: Rethinking model scaling for convolutional
  neural networks,'' 2020.

\bibitem{dosovitskiy2021image}
A.~Dosovitskiy, L.~Beyer, A.~Kolesnikov, D.~Weissenborn, X.~Zhai,
  T.~Unterthiner, M.~Dehghani, M.~Minderer, G.~Heigold, S.~Gelly, J.~Uszkoreit,
  and N.~Houlsby, ``An image is worth 16x16 words: Transformers for image
  recognition at scale,'' 2021.

\end{thebibliography}
\end{document}